# TAGE: Trustworthy Attribute Group Editing for Stable Few-shot Image Generation

Ruicheng Zhang1, Guoheng Huang1*, Yejing Huo1, Xiaochen Yuan2,

Zhizhen Zhou1, Xuhang Chen3, Guo Zhong4*

1School of Computer Science and Technology, Guangdong University of Technology,

Guangzhou, 510006, China.

2Faculty of Applied Sciences, Macao Polytechnic University, Macao, China.

3School of Computer Science and Engineering, Huizhou University, Huizhou, 516001,

China.

4School of Information Science and Technology, Guangdong University of Foreign Studies,

Guangzhou, 510006, China.

## ABSTRACT

Generative Adversarial Networks (GANs) have emerged as a prominent research focus for image editing tasks, leveraging the powerful image generation capabilities of the GAN framework to produce remarkable results. However, prevailing approaches are contingent upon extensive training datasets and explicit supervision, presenting a significant challenge in manipulating the diverse attributes of new image classes with limited sample availability. To surmount this hurdle, we introduce TAGE, an innovative image generation network comprising three integral modules: the Codebook Learning Module (CLM), the Code Prediction Module (CPM) and the Prompt-driven Semantic Module (PSM). The CPM module delves into the semantic dimensions of category-agnostic attributes, encapsulating them within a discrete codebook. This module is predicated on the concept that images are assemblages of attributes, and thus, by editing these category-independent attributes, it is theoretically possible to generate images from unseen categories. Subsequently, the CPM module facilitates naturalistic image editing by predicting indices of category-independent attribute vectors within the codebook. Additionally, the PSM module generates semantic cues that are seamlessly integrated into the Transformer architecture of the CPM, enhancing the model's comprehension of the targeted attributes for editing. With these semantic cues, the model can generate images that accentuate desired attributes more prominently while maintaining the integrity of the original category, even with a limited number of samples. We have conducted extensive experiments utilizing the Animal Faces, Flowers, and VGGFaces datasets. The results of these experiments demonstrate that our proposed method not only achieves superior performance but also exhibits a high degree of stability when compared to other few-shot image generation techniques.

## 1. INTRODUCTION

Few-Shot Image Generation [8, 37, 6] is an important research direction in the field of computer vision and deep learning, and its goal is precisely to learn the potential patterns captured from a very limited number of examples, and utilize this information to generate diversified and realistic new images of the corresponding categories or other unseen categories. This technique is particularly suitable for those cases where only a small amount of labeled or exemplar data is available, and is important for solving the problems of model generalization ability and adaptability when there is insufficient data. In practical applications, such as personalization [24], art creation [23], medical image analysis [19, 5,39], image enhancement [26, 25, 4] and dealing with rare category object recognition, few-sample image generation shows great potential and value.

Current few-shot image generation methods are broadly classified into three categories: optimization-based, fusion-based, and transformation-based. Optimization based approaches [6, 21] learn a set of generalized basic models through meta-learning and fine tunes them for different tasks to achieve the goal. However, the quality of images generated by these methods is not high. Fusion based methods [17, 16, 12] extract features from different input images and fuse them into new categories of images in the latent space, the limitation of this method is that the input images need to be relatively similar and are quantitatively demanding. Transformation based methods [32] want to find intra-category transformations and apply these transformations to unseen category samples to generate more images of the same category, the disadvantage is that the transformations are complex and training is unstable.

In contrast to the previous three methods, editing-based methods model the generation of few-shot images as attribute editing problems, which allows us to avoid complex and unstable transformation structures during training, and achieve high-quality image generation. The first edit-based approach proposed is AGE [8], which has high production quality. However, there are still many problems. As shown in Figure 1, the images generated by AGE may result in the disappearance and collapse of organs, significantly impacting the perceived image quality. In pursuit of the primary objective to generate training data for a handful of downstream applications, the quality improvements achieved thus far do not meet the expected standards of satisfaction. In this paper, we further study the generation of new category images based on codebook and text prompts, with the goal of enhancing the model's stability and attribute editing controllability.

In this paper, we propose TAGE, an image generation model that leverages pre-trained Generative Adversarial Networks (GANs) to mine semantic directions and perform attribute editing without direct supervision. TAGE introduces dictionary learning into few-shot image generation using three key modules: the Codebook Learning Module (CLM), Code Prediction Module (CPM), and Prompt-driven Semantic Module (PSM). The CLM uses unlabeled images to identify semantic directions for both category-related and unrelated attributes, constructing a sparse dictionary for generating unseen category images by recombining known attributes. The CPM enhances control and stability by predicting latent codes that ensure accurate attribute editing, even under limited data or high diversity conditions. The PSM generates semantic prompts that guide the CPM, enabling fine-grained control over attribute manipulation while preserving coherence. Together, these modules allow TAGE to extract semantic information from pre-trained GANs and achieve flexible, high-quality image generation and editing, particularly in few-shot scenarios.

Our contributions can be summarized as follows:

1. We propose a few-shot image generation method called TAGE, including Codebook Learning Module (CLM), Code Prediction Module (CPM) and Prompt-driven Semantic Module (PSM). The method identifies category-independent editing directions without explicit supervision and enables more stable attribute editing.

2. In few-shot image generation scenarios, a limited small-scale potential space helps to improve image quality. Our proposed CLM achieves this by limiting the potential space and storing high-quality reconstruction elements.

3. To address the dilemma of lower input quality and reduced diversity, our proposed CPM enables better code prediction using global combinatorial information and long-range dependencies to improve the diversity of the generated images.

4. The Prompt-driven Semantic Module (PSM) facilitates few-shot learning by injecting semantically-guided prompts into the transformer layers, enabling better attribute understanding and manipulation for higher-quality image generation and editing with scarce data.

5. The experimental results from the Animal Faces, Flowers, and VGG Faces datasets demonstrate that our proposed network can generate higher quality images with notable improvements in performance.

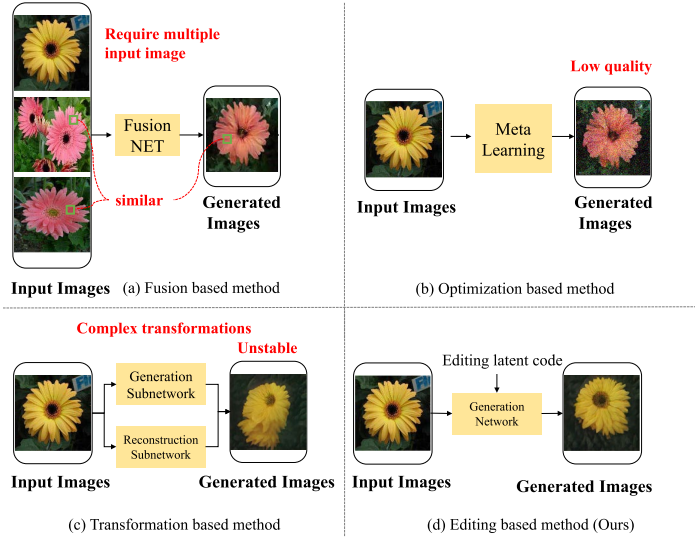

Figure 1. The comparison of four existing methods on few-shot image generation. (a) Fusion-based methods [17, 16, 12]: They involve the fusion of features from multiple input images in latent space to generate new image categories. The similarity constraint of the input can limit the creativity and variety of the resulting images, potentially leading to a lack of generalisation. (b) Optimization-based methods [6, 21]: Despite their ability to learn generalized models through meta-learning. The optimization process is not sufficient to capture the intricate details of the unseen category, leading to images that are less realistic. (c) Transformation-based methods [32]: The complexity of learning and applying transformation patterns within the same category can be a significant challenge for these methods, the transformations may not always be accurately learned or applied. (d) Editing-based methods [8]: They avoid complex and unstable transformation structures during training, and achieve high- quality image generation.

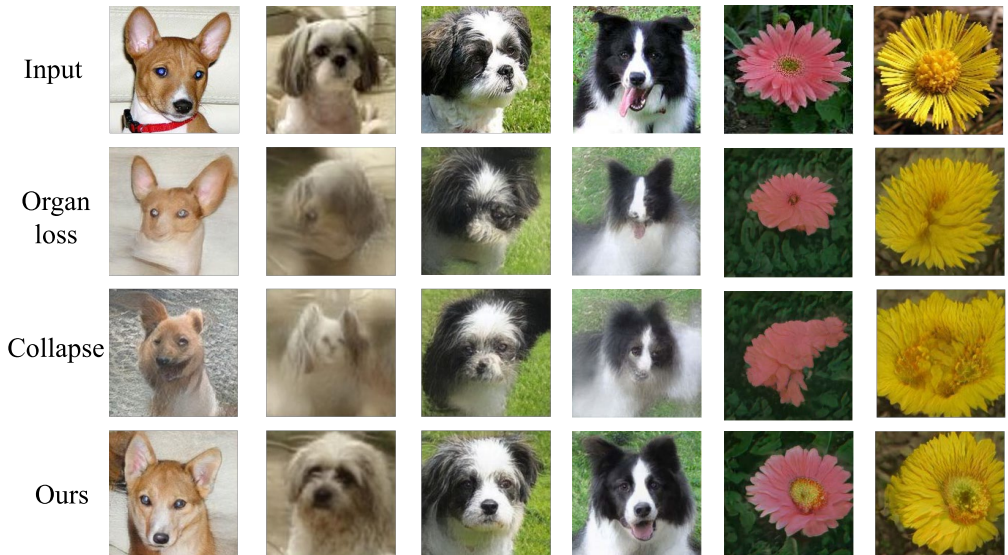

Figure 2. Image crash phenomenon in AGE [8]. AGE suffers from two major limitations. Firstly, due to its sampling during the inference process being based on the statistical data of the training set rather than being adaptive to the input images, it can result in generated objects appearing in irregular poses. Secondly, while some attributes are not irrelevant to all categories, they are acquired in a way that's not tied to a specific category. This results in generated images that, despite their realism, experience a shift in category. If the category-relevant attributes of the input image cannot be embedded effectively, editing will also fail.

## 2. RELATED WORK

### 2.1 Few-shot Image Generation

As shown in Figure 0, few-shot image generation research can be grouped into three paradigms: optimization-based, fusion-based, and transformation-based methods. Optimization-based methods [6, 21] use meta-learning to train generalized base models that are fine-tuned for different tasks, but they often generate lower-quality images due to insufficient detail capture. Fusion-based methods [17, 16, 12] combine features from multiple input images in the latent space to create new image classes, yet they rely on high similarity among inputs and are computationally expensive. Transformation-based approaches [32] apply intra-class variations to unseen categories to synthesize images, but their complexity and instability during training are significant limitations.

### 2.2 Codebook Learning

Sparse dictionaries have proven effective in image tasks like super-resolution and denoising. The VQ-VAE framework [30, 11] learns discrete codebooks in latent space, addressing "posterior collapse" and enhancing model performance. VQGAN [10] further improves perceptual quality through adversarial training, while CodeFormer [38] replaces Nearest-Neighbor Matching with a Transformer-based network for better codebook prediction. Leveraging these state-of-the-art methods, we use discrete codebooks for few-shot image generation to improve image quality and robustness.

### 2.3 Text-driven Image Generation

Initially dominated by GANs [36, 33], text-driven image generation has shifted toward diffusion models [2, 18], which integrate advanced text processing for more precise image synthesis. For instance, DAELL2 [31] and StyleCLIP [29] combine CLIP embeddings with image generation models for high-fidelity results. Transformer-based models like CogView2 [9] and Muse [3] have also shown strong performance. Unlike these methods, our approach performs unsupervised semantic editing in StyleGAN's latent space for few-shot image generation, incorporating prior text embeddings to enhance image quality without requiring extensive labeled data or complex loss functions.

## 3. THE PROPOSED METHOD

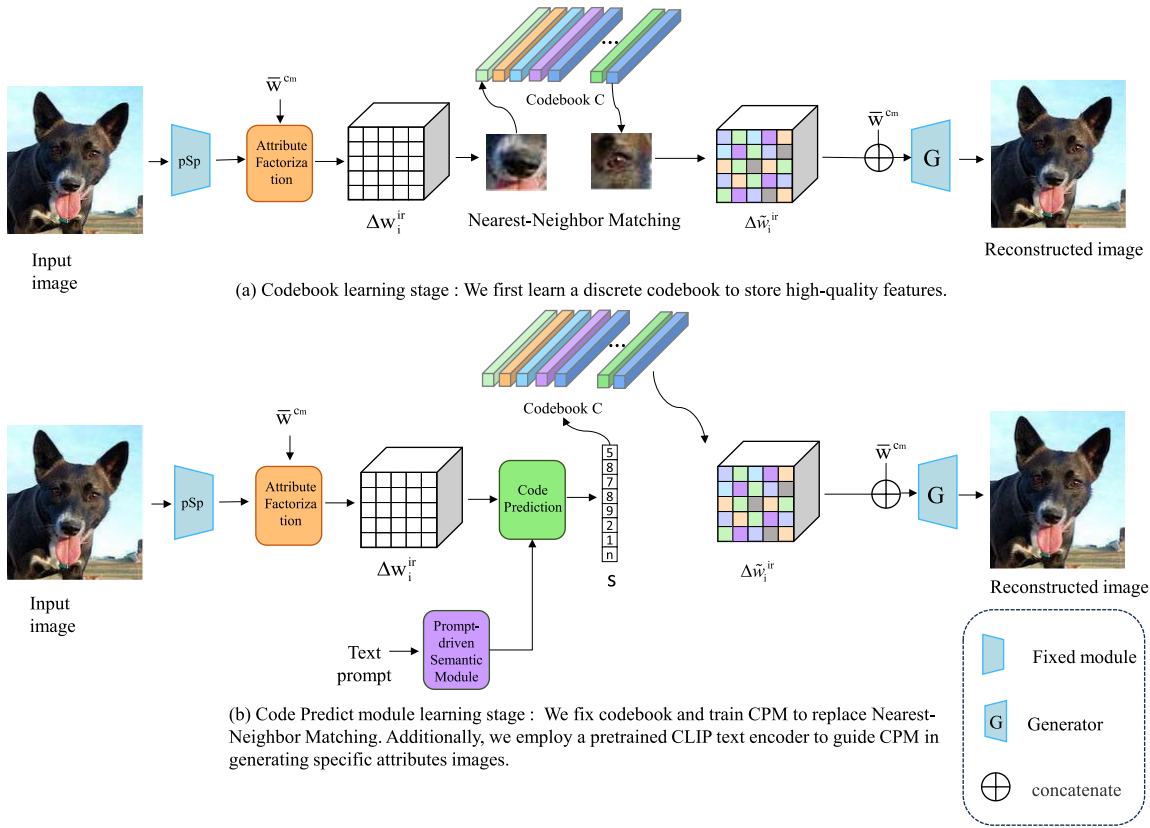

(a) Codebook learning stage : We first learn a discrete codebook to store high-quality features.

(b) Code Predict module learning stage : We fix codebook and train CPM to replace Nearest-Neighbor Matching. Additionally, we employ a pretrained CLIP text encoder to guide CPM in generating specific attributes images.

Figure 3. The illustration of our proposed model. Our training steps consists of two steps, the specific compositions are showed in (a) and (b). In stage (a), the image is encoded by a model named "pSp" to obtain a latent representation $w^{cm}$. Then the Attribute Factorization Module extract category-irrelevant attribute $\Delta w_i^{ir}$ from the latent representation. This category-irrelevant attribute is further processed to extract and store high-quality features in a discrete codebook $C$. The codebook is created to capture the semantic directions of category-irrelevant attributes, which are essential for editing without explicit supervision. In stage (b), the fixed codebook from the first stage is used, and the Code Prediction Module (CPM) is trained to improve

the prediction of the code sequence that will be used to generate the image. Additionally, a pre-trained CLIP text encoder is employed to guide the CPM in generating images with specific attributes.

### 3.1 The overview of our method

We use the seen category $c_s$ as training set and unseen category $c_u$ as testing set, where the number of images in $c_u$ is small. Our goal is to generate unseen category images by editing category-irrelevant attributes, whose direction is extracted from a large number of seen category images without explicit supervision. Our method framework is shown in Figure 3. In the training stage, we embed the image into latent space and distinguish category-irrelevant attributes from category-relevant attribute vectors. Inspired by the idea of dictionary learning [13], we use CLM to discretize and store the semantic directions of category-irrelevant attributes in a dictionary model, and employ a Code Prediction Module and a Prompt-driven Semantic Module to predict the code combination to achieve stable attribute group editing.

### 3.2 Attribute Factorization

To realize attribute editing, the real image has to be firstly mapped to the latent space. This step is very important, because the performance of the image editing largely relies on the quality of the latent code. We use pSp as our encoder to find the latent code of real images in the latent domain. pSp use feature pyramid as backbone to encode image into three levels feature maps, which correspond to the coarse, medium and fine details in StyleGAN.

$$w_i = pSp(x\_i) \tag{1}$$

where the $x_i$ is the input image and $w_i \in \mathbb{R}^{18 \times 512}$ is the latent code that corresponds to input image.

Assuming we already have latent code in $w+$ space, the next thing to do is to separate a set of category-relevant attribute and category-irrelevant attribute directions. As mentioned above, theoretically a large number of images of unseen categories can be generated by editing category-irrelevant attributes. But without explicit supervision finding the category-irrelevant attribute directions is difficult. Since the category-related vectors of the same category are similar, if a large number of embedding vectors $w_i^{cm}$ of category $c_m$ are given, their average vectors approximate the category-related vectors $\overline{w}^{cm}$ we need.

$$\overline{w}^{cm} = \frac{1}{N_m} \sum_{i=1}^{N_m} w_i^{cm} \tag{2}$$

where $N_m$ is the number of samples in category $c_m$. Then, for a seen category, the latent code of image can be composed of the category-relevant vectors $\overline{w}^{cm}$ plus the category-irrelevant attribute vectors $\Delta w_i^{ir}$ that we want to obtain.

$$w_i^{cm} = \overline{w}^{cm} + \Delta w_i^{ir} \tag{3}$$

With this algorithm we are able to obtain a large number of category-irrelevant attribute vector $\Delta w_i^{ir}$, which provides a stable data source for our next step training. To increases the diversity and stability of attribute editing, we want to employ a Code Prediction Module to predict the category-irrelevant attribute vector. We first incorporate the idea of dictionary learning, using a pre-trained encoder to obtain a discrete codebook. Our method's training is divided into two stages accordingly.

### 3.3 Codebook Learning Module

The goal of the first stage of training is to train a context-rich codebook [20], given a category-irrelevant attribute vector, we joint optimize a global dictionary $A \in \mathbb{R}^{18 \times 512 \times l}$ which contains category-irrelevant directions and a sparse representation $n_i$ following the AGE training.

Then replace each "pixel" of the reconstructed category-irrelevant vector $\Delta w_i^{ir}$ that generated by dictionary $A \in \mathbb{R}^{18 \times 512 \times l}$ with the closest part in the codebook $C \in \{c_k \in \mathbb{R}^{512}\}_{k=0}^{N}$ to obtain a new reconstructed category-irrelevant vector $\Delta \widetilde{w}_i^{ir}$.

$$\Delta \widetilde{w}_i^{ir(i,j)} = \arg\min \| \Delta w_i^{ir(i,j)} - c_k \|_2 \tag{4}$$

The reason why we still need a discrete codebook when we already have a global dictionary A is that compare to the continuous infinite space, small finite proxy space shows superior robustness and reconstruct quality. When input an unseen category image, the embedding modules such as pSp probably generate an ambiguous latent code, which will seriously affect the subsequent of the following editing and generation. The discrete structure force codebook to conserve high quality details, and low-quality latent code has higher probability to match the accurate code in codebook.

To train the context-rich codebook, we adopt two losses:

The $L_2$ reconstruction loss is to optimize the generated images close to the input image.

$$L_{rec} = \| G(\overline{w}^{cm} + \Delta w_i^{ir}) - x_i^{c_m} \|_2 \tag{5}$$

The codebook loss is to reduce the distance between codebook C and the embedding of the edited image.

$$L_{codebook} = \| sg(\Delta w_i^{ir}) - \Delta \tilde{w}_i^{ir} \|_2 + \beta \| \Delta \tilde{w}_i^{ir} - sg(\Delta w_i^{ir}) \|_2 \tag{6}$$

where sg(·) stands for the stop-gradient operator and $\beta$ is a hyper parameters that control the update rates of the dictionary and codebook.

### 3.4 Code Prediction Module

Since the unseen categories of images are varied, sometimes Nearest-Neighbor(NN) Matching usually fails to find the accurate editing code, making the edited images have a lower degree of perceived quality. Besides, diversity is important To alleviate this problem, we employ a Transformer module and some linear layers to predict the category-irrelevant attribute vector. we insert a Transformer [34] module which contains nine self-attention blocks following the dictionary module. The structure is shown in Figure 1. At this stage we freeze all modules except the Code Prediction Module, The i-th self-attention block of Transformer computes as the following:

$$X_{i+1} = \sigma(Q_i K_i) V_i + X_i \tag{7}$$

where the $X_0 = \Delta w_i^{ir}$, The query Q, key K, and value V are obtained from $X_i$ through linear layers.

Generally speaking, CPM use the edited image vector $\hat{w}_i^{ir}$ as an input to predict n layer code sequence $s \in \{0, \cdots, N-1\}^n$ represent the probability of the N code items. Then, based on the predicted code sequence s, n individual code items are retrieved from the codebook to form the reconstruct vector.

In stage II, since we only train CPM, we don't need the four losses mentioned above. We only need two code-level losses: 1) cross-entropy loss $L_{cross-entropy}$ for code prediction supervision, and 2) $L_2$ loss $L_{code}^{dict}$ for the prediction code $\Delta \tilde{w}_i^{ir}$ close to the embedding of the edited image $\Delta w_i^{ir}$.

$$X_{i+1} = \sigma(Q_i K_i) V_i + X_i \tag{8}$$

where $s_i$ is the ground truth code sequence s is obtained from the stage I and $\tilde{s}_i$ is predicted, n represent the pixel number in $\Delta w_i^{ir}$.

$$L_{code}^{dict} = \| \Delta w_i^{ir} - sg(\Delta \tilde{w}_i^{ir}) \|_2 \tag{9}$$

The overall loss function is:

$$L_{tf} = L_{cross-entropy} + L_{code}^{dict} \tag{10}$$

In the inference phase, the same as AGE, we sample an arbitrary $\hat{n}_j$ from $N(\mu, \Sigma)$ and apply editing to unseen category images. When we get the embedding of the edited image, we put it to the Code Prediction Module to predict the code sequence and generate the image like stage II.

### 3.5 Prompt-driven Semantic Module

In this module, we first need to define the structure of cue words, a system of cue words containing category information, color features, shape features, and environment or background elements. Second, we need to construct a vocabulary V that covers all predefined cue words. Next, we utilize the text encoder in CLIP to extract word embedding vector $\mathbf{v}_i = E^{CLIP}(t_i)$ for each word $t_i \in V$ in the glossary. Where $E^{CLIP}$ denotes the pre-trained CLIP image encoder.

For the input image, the model randomly selects a set of relevant cue words $\{t_1, t_2, \ldots, t_n\}$ from the vocabulary list. Each word in the cue word sequence is transformed into a corresponding word embedding vector to form a cue word vectors $\mathbf{v_i}$. The Transformer structure accepts both a sequence of edited image vector $\Delta \tilde{w}_i^{ir}$ and the cue word embedding vectors $\mathbf{v_i}$ for interactive computation, with the following formula:

$$Q_i = \tilde{w}_i^{ir} W_Q \tag{11}$$

$$K_i = v_i W_K \tag{12}$$
$$V_i = v_i W_v \tag{13}$$
$$X_{i+1} = \sigma(Q_i K_i) V_i + X_i \tag{14}$$

To make sure the subject consistency between the input image and output images, we feed then through CLIP encoder to check the distance between these two embeddings. If the two images have the same subject, the distance should be small. The subject loss function is defined as follows:

$$L_{sub} = 1 - \cos\{E_{CLIP}(G(w')), E_{CLIP}(G(w))\} \tag{15}$$

where the $\cos\{\cdot\}$ stands for cosine similarity, $w'$ and $w$ are the edited and input latent code respectively.

To measure the correlation between output images and cue word embeddings, we minimize their cosine distance of the CLIP embeddings. The word loss function is defined as follows:

$$L_w = 1 - \cos\{E_{CLIP}(G(w_i')), v_i\} \tag{16}$$

Through this interactive work, the image features are able to receive guidance from the cue word information during the encoding process, which facilitates the generation of enhanced specific attributes.

## 4. EXPERIMENT

### 4.1 Implementation Details

We use pre-trained StyleGAN generator and pre-trained pSp encoder. The MLP is 5 layers with Leaky-ReLU activation function. We set the length $l$ of dictionary A to 100 and the codebook size N to 10000. Our method is trained using two NVIDIA RTX 3090 GPUs with the PyTorch framework.

### 4.2 Datasets

We conduct experiment on three few-shot image datasets: Animal Faces [22], Flowers [27], VGGFaces [28], each dataset is split into two parts: seen category for training and unseen category for testing.

Animal Faces: The dataset consists of images of 149 carnivore categories from ImageNet [7]. The dataset contains 117,574 carnivore images. We divided these classes into a source class set and a target class set, containing 119 and 30 animal classes, respectively.

Flowers: The dataset is an image classification dataset mainly used to test the performance of the algorithm in complex scenarios. It contains 102 different flower classes, which are mainly some common species in the UK. Each category contains images ranging from 40 to 258 images, totalling 8189 images.

VGGFaces: The dataset contains 3.31 million images from 9,131 celebrities spanning a wide range of races and professions. The dataset is divided into two parts: one for training with 1802 classes and the other for evaluation (testing) with 552 classes.

### 4.3 Metric

We evaluate our method by FID [14] and LPIPS [35] metrics. FID is a commonly used metric for evaluating the generative model, it measures the performance of the generative model by comparing the distance between the distribution of the generated image and the distribution of the real image, if the value of FID is smaller, it means that the distance between the generative model and the real distribution is smaller, and the quality of generation is better. LPIPS is used to evaluate the perceptual similarity to measure the similarity of images, it can be used to measure the image with similarity by simulating the human eye's perception of the picture, we adopt LPIPS to measure the diversity of generated images.

### 4.4 Quantitative Evaluation

We compare our network with other few-shot image generation methods in Animal Faces, Flowers, VGGFaces datasets. We randomly select from each unseen category image and generate 128 fake images, which are denote as $\mathbb{S}_{fake}$. And extract the equal number images from each unseen category image as $\mathbb{S}_{real}$. We calculate the FID between $\mathbb{S}_{real}$ and $\mathbb{S}_{fake}$, and only use $\mathbb{S}_{fake}$ to calculate the LPIPS score, the result of different methods is showed in Table 1. Our model has achieved some improvements in both FID and LPIPS. We achieved the best LPIPS scores on both datasets and top two on FID. And our method requires only one image as input and produces more diverse images. Compared to AGE, our method is more stable, as will be illustrated in the qualitative evaluations below.

| Methods | k-shot | Flowers | | Animal Faces | | VGG Faces | |
|---|---|---|---|---|---|---|---|
| | | FID ↓ | LPIPS ↑ | FID ↓ | LPIPS ↑ | FID ↓ | LPIPS ↑ |
| FIGR [6] | 3 | 190.12 | 0.0634 | 211.54 | 0.0756 | 139.83 | 0.0834 |
| GMN [1] | 3 | 200.11 | 0.0743 | 220.45 | 0.0868 | 136.21 | 0.0902 |
| DAWSON [21] | 3 | 188.96 | 0.0583 | 208.68 | 0.0642 | 137.82 | 0.0769 |
| DAGAN [32] | 1 | 179.59 | 0.0496 | 185.54 | 0.0687 | 134.28 | 0.0608 |
| MatchingGAN [17] | 3 | 143.35 | 0.1627 | 148.52 | 0.1514 | 118.62 | 0.1695 |
| F2GAN [16] | 3 | 120.48 | 0.2172 | 117.74 | 0.1831 | 109.16 | 0.2125 |
| LofGAN [12] | 3 | 79.33 | 0.3862 | 112.81 | 0.4964 | **20.31** | 0.2869 |
| DeltaGAN [15] | 3 | 109.78 | 0.3912 | 89.1 | 0.4418 | 80.12 | 0.3146 |
| AGE [8]* | 1 | 90.03 | 0.4365 | **61.67** | 0.5411 | 34.86 | **0.3156** |
| Ours | 1 | **79.21** | **0.4403** | 70.13 | **0.5582** | 34.78 | 0.3021 |

Table 1. Comparison of results of different methods. The best performances are presented in bold, and the second best are showed in blue. AGE is marked with * because our dataset was obtained from LoFGAN, and the dataset partitioning used by AGE has not been published, so the metrics are different from the original paper.

### 4.5 Qualitative Evaluation

In qualitative evaluation, we also compare with the one-shot image generation method AGE on three datasets: Animal Faces, Flower, VGGFaces. It can be seen that both AGE and TAGE have some generalization ability to generate images with different attributes. For example, we can generate dogs and flowers with different positions, man with sad or smile expression.

However, the indicators of a successful redevelopment are not just about diversity, perceptual is also a important metric. Imagine we want to reconstruct a picture of a person, one picture perfectly reconstructs all the details except for a missing nostril, and the other one may be missing some details but is complete, which one would be better? We believe is the latter one, because it looks more like real picture. Compare with AGE, the images generated by TAGE are more in accord with human perception compared to AGE. For instance, like Figure 4, the images generated in AGE sometimes lose some of the organs such as the eyes and nose, causing the image to collapse. However, this is rarely the case in TAGE, thanks to the discrete codebook structure, Code Prediction Module and Prompt-driven Semantic Module, which makes the perceptual quality of TAGE generated images higher.

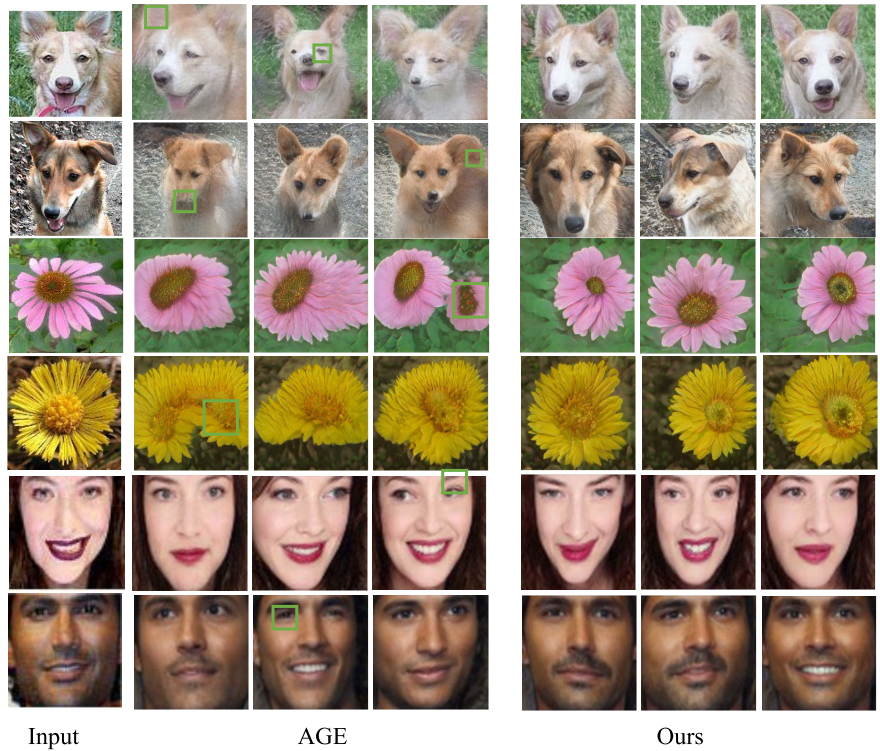

Input      AGE      Ours

Figure 4. Comparison between images generated by AGE [8] and our method on Flowers, Animal Faces, and VGGFaces. As can be seen in the figure, there are some animal organ loss and partial image collapse phenomena in the images generated by the AGE method. Such as blurred noses and flowers collapsing in two.

## 4.6 Ablation Study

In order to verify the effectiveness of the proposed module, we perform ablation experiments on Animal Faces and Flowers dataset. The result is shown in Table 2 and figure 5.

1) Importance of Codebook Module: A codebook can be considered as a discrete dictionary, the discretization is done to force the separation of different attributes, the disentanglement of attributes is very important to improve the quality of image editing. The images generated only by the codebook can store more details such as the head and ear colour, which makes it perform better in the FID score. On the other hand, although the FID score is high, the images generated only by codebook are similar to the input images. For example, the dogs' position is same and mouth open similar way, which means they lack of diversity. These results show that when the dictionary become discrete, the quality of the image editing is improved and the diversity is falling. This is also foreseeable because compare to the continuous dictionary, the discrete codebook is more concentrate on the optimal elements for high quality image reconstruction. For the task of few-shot image generation, we need to combine the Code Prediction Module to improve the diversity of generation, and it is not meaningful to use the codebook alone.

2) Important of Code Prediction Module: To enhance the diversity of image editing and reduce image collapse, we design a Code Prediction Module to replace the Nearest-Neighbour(NN) matching. The Code Prediction Module is composed of liner layer and Transformer, aiming to predict better embedding of the edited image. In Figure 5, we can see the diversity is much higher than generated only by codebook, and still retain the input images' features such as ears and mouth shape. As the Table 2 shows, our FID down to 74.41 and LPIPS improve to 0.5501. Most importantly, the organs missing phenomenon sharply decrease with the Code Prediction Module, which is necessary for a stable image editing.

3) Important of Prompt-driven Semantic Module: To improve model understanding and precise attribute manipulation, we introduce the Semantic Prompt Module (PSM). PSM generates prompts injected into the Code Prediction Module's (CPM) Transformer layers. Images with PsM show more prominent and coherent editing of target attributes. In animal face editing, PSM preserves species traits while accentuating edited attributes like age or expression. As the Table 2 shows, our FID improves to 70.13 and LPIPS improve to 0.5582. PSM addresses attribute distortion or missing regions in complex visual editing, ensuring coherent structures and faithful attribute modifications even with limited data.

| Network | | | Animal Faces | | Flowers | |
|---|---|---|---|---|---|---|
| CLM | CPM | PSM | FID ↓ | LPIPS ↑ | FID ↓ | LPIPS ↑ |
| √ | | | 70.25 | 0.5206 | 79.67 | 0.3478 |
| √ | √ | | 74.41 | 0.5501 | 82.21 | 0.4396 |
| √ | √ | √ | **70.13** | **0.5582** | **79.21** | **0.4403** |

Table 2. Ablation study. The CLM enhances the quality of image editing by discretizing dictionaries, albeit at the cost of reducing diversity. Conversely, the introduction of the CPM not only augments the diversity of image editing but also mitigates image collapse by predicting superior quality embedding codes. The inclusion of the PSM module generates pictures of unchanged kinds with more obvious attributes while ensuring that the model better understands the attributes to be edited. Note: The best performances are presented in bold.

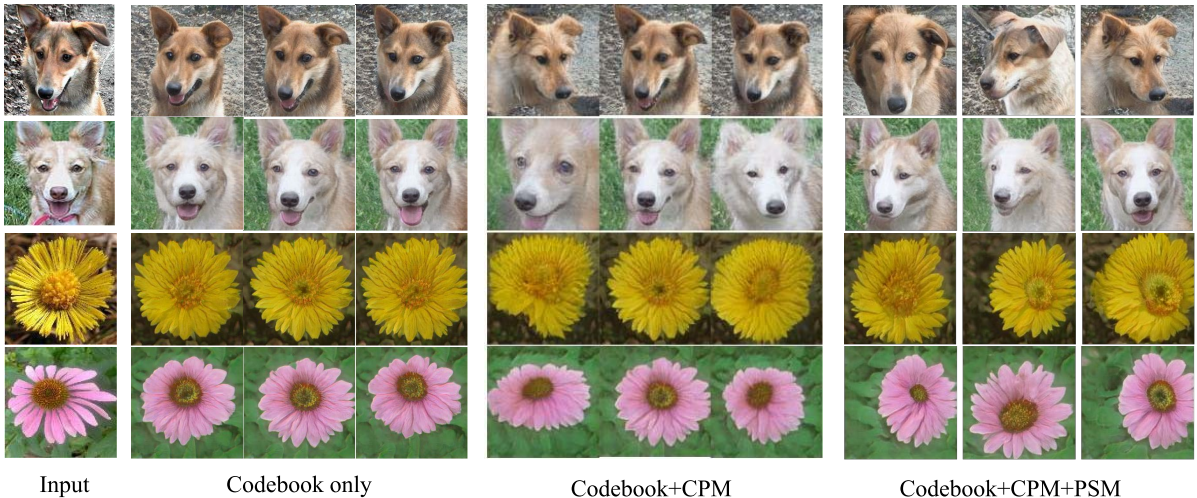

Input     Codebook only     Codebook+CPM     Codebook+CPM+PSM

Figure 5.   Visualization of ablation study.

### 4.7 User Study

To further verify the stability and the realism of our generated images, we conduct a user study for our method. In addition to our method, we chose AGE that has excellent performance as our opponent. We randomly select three images of unseen categories from different validation datasets as the evaluation set. For each input image, we separately generated three edited images as a set using AGE and TAGE. A total of 50 participants was then invited to discern and choose the set of images that they perceived as appearing more natural following the editing process. As suggested in Figure 6, TAGE demonstrated superior performance in terms of both perceptual quality and stability. This signifies that the images generated by TAGE were deemed more visually appealing and stable by the study participants.

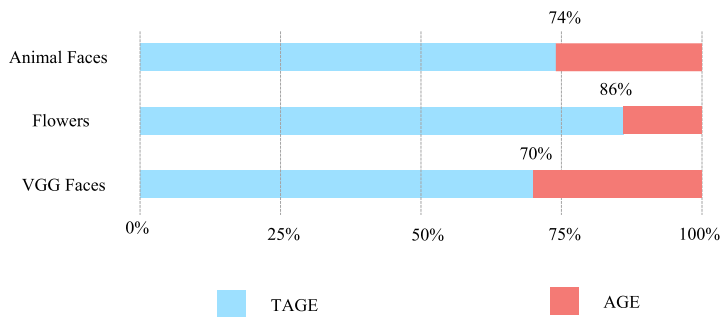

Figure 6.   Result of user study. Voting statistics of AGE versus our method.

## 5. CONCLUSION

We propose Trustworthy Attribute Group Editing (TAGE), a novel method for unsupervised attribute group editing that mitigates the crash phenomenon seen in attribute group editing. TAGE comprises three key components: (1) the Codebook Learning Module (CLM) that learns a discrete codebook to constrain the latent space and improve semantic direction for better image generation, (2) the Code Prediction Module (CPM) that replaces Nearest-Neighbor matching to enhance diversity and prevent crashes caused by low-quality latent codes, and (3) the Prompt-driven Semantic Module (PSM) that allows the model to understand target attributes more effectively, enabling clearer attribute generation with fewer samples. Extensive experiments demonstrate TAGE's stability and diversity in few-shot image generation. While TAGE excels in many areas, some limitations remain. Occasionally, generated images may shift categories, such as changes in a dog's color or petal count affecting the perceived category. Future research will focus on advanced strategies to disentangle category-irrelevant attributes to minimize these unintended shifts.

## 6.   ACKNOWLEDGEMENT

This work was supported by Key Areas Research and Development Program of Guangzhou Grant 2023B01J0029, Science and technology research in key areas in Foshan under Grant 2020001006832, the Science and technology projects of Guangzhou under Grant 202007040006, the Guangdong Provincial Key Laboratory of Cyber-Physical System under Grant 2020B1212060069.